\documentclass[11pt,a4paper]{article}

\usepackage[table]{xcolor}
\usepackage{url}
\usepackage[hyperref]{acl2018}
\usepackage{times}
\usepackage{latexsym}
\usepackage{threeparttable}
\usepackage{graphicx}
\usepackage{paralist}
\usepackage{enumitem}
\usepackage{breakcites}

\usepackage[font=small,format=plain,justification=justified,singlelinecheck=false,labelfont=bf,up,textfont=small,up]{caption}

\aclfinalcopy 

\title{Detecting Adverse Drug Reactions from Twitter through Domain-Specific Preprocessing and BERT Ensembling}

\author{Amy Breden \\
  UC Berkeley School of Information\\
  {\tt amy.breden@berkeley.edu} \\\And
  Lee Moore \\
  UC Berkeley School of Information\\
  {\tt lee.moore@berkeley.edu} \\}

\date{}

\begin{document}
\maketitle

\begin{abstract}
The automation of adverse drug reaction (ADR) detection in social media would revolutionize the practice of pharmacovigilance, supporting drug regulators, the pharmaceutical industry and the general public in ensuring the safety of the drugs prescribed in daily practice. Following from the published proceedings of the Social Media Mining for Health (SMM4H) Applications Workshop \& Shared Task in August 2019, we aimed to develop a deep learning model to classify ADRs within Twitter tweets that contain drug mentions. Our approach involved fine-tuning BERT\textsubscript{LARGE} and two domain-specific BERT implementations, BioBERT and Bio + clinicalBERT, applying a domain-specific preprocessor, and developing a max-prediction ensembling approach. Our final model resulted in state-of-the-art performance on both $F_1$-score (0.6681) and recall (0.7700) outperforming all models submitted in SMM4H 2019 and during post-evaluation to date.  
\end{abstract}

\section{Introduction}

Adverse drug reactions are one of the leading causes of hospitalizations and deaths globally. New pharmaceuticals are approved by the FDA with limited clinical data relative to the size of the populations subsequently prescribed these drugs, making their real-world safety profile uncertain. Current methods for ADR monitoring such as post-marketing surveillance studies do not take advantage of the wealth of information shared by the public on the internet, resulting in an untapped opportunity for automatic detection and extraction of this information from social media platforms such as Twitter. Existing models underperform in this context due to misspellings and frequent use of idiomatic, ambiguous and sarcastic expressions in social media \cite{Sarker:2015}.  
	
When considering evaluation criteria it is important to take into account the context of the potential use case. The value of automatic ADR detection in social media for pharmacovigilance will be for (1) identifying novel ADRs in a wider population not currently identified by more traditional data collection means and (2) increasing the speed by which ADRs are identified. In the context of Twitter, drug use mentions are rare with less than 1\% of all tweets even mentioning drug names \cite{Meng}, with our dataset further highlighting that less than 10\% of these drug-related tweets may represent ADR mentions. For these reasons, we argue that recall is the most important performance metric given the focus on identifying all true ADRs. Whilst precision is relevant, it is not critical since false postive ADRs can be filtered out in downstream tasks, so there is limited harm in falsely identified ADRs while there is significant public good in correctly identifying ADRs which may not be identified as readily using current methods.
	
Our study aims to achieve state-of-the-art results, specifically on $F_1$-score and recall, on a dataset used as part of an annual competition on classifying Twitter tweets as either reporting an adverse drug reaction or not. We use  BERT\textsubscript{LARGE} as our baseline, with hyperparameters selected based on the 2019 winning team’s model. For our contribution, we fine-tuned two domain-specific BERT implementations, BioBERT \cite{biobert} and Bio + clinicalBERT \cite{alsentzer-etal-2019-publicly}, applied a domain-specific preprocessor, and developed a novel ensembling approach. 

\begin{table*}[t!]
\small 
\caption{\label{preprocessor} Preprocessor Components and Description}
\begin{tabular}{ |l|l|l| } 
 \hline
  \bf Component & \bf \bf Description\\
 \hline
  Anonymize & Replaces URLs with "-URL-", removes usernames from email addresses while retaining the email\\
   &  domain, and removes copyright and trademark logos \\
  Replace handles & Replaces twitter handles with “-TH-” \\
  Remove hashtags & If word starts with “\#”, remove “\#” symbol but keep the word\\
  Lowercase & Converts tokens to lowercase (required for Drug Normalizer step)\\
  Drug normalize & Converts drug brand names to generic names\\
 \hline
\end{tabular}
\end{table*}

\section{Background}

The first notable attempts at classifying ADRs from Twitter were published in 2015 \cite{Sarker:2015} applying natural language processing techniques for feature extraction and traditional machine learning models such as support vector machine classifiers. Following this, an annual meeting entitled Social Media Mining for Health (SMM4H) Applications Workshop \& Shared Task arose which has met over four consecutive years, with the latest meeting in August 2019. An annotated corpus of tweets has developed alongside this work, currently consisting of 25,678 labeled tweets available for training and an additional 4,575 tweets used for model evaluation as part of the ongoing competition \cite{weissenbacher-etal-2019-overview}.
	
The most recent meeting of SMM4H confirmed that techniques used to tackle this task have evolved alongside the developments in deep learning architectures more generally, with the 2018 dominance of convolutional and recurrent neural networks being overtaken in 2019 with neural architectures using word embeddings pretrained with Bidirectional Encoder Representations from Transformers (BERT) \cite{devlin-etal-2019-bert}. The performance on classifying tweets improved from 0.522 $F_1$-score (0.442 precision, 0.636 recall) in SMM4H 2018 to 0.646 $F_1$-score (0.608 precision, 0.689 recall) in SMM4H 2019 \cite{weissenbacher-etal-2019-overview}. The winning team’s approach involved retraining BERT on a large-scale unlabeled corpus of 1.5 million tweets extracted from Twitter based on the 150 drug names collected from the training set \cite{chen-etal-2019-hitsz}. While these are the highest scores seen to date on this dataset, there is still room for improvement on this task. One such improvement of the BERT-based models presented at SMM4H 2019 could be the inclusion of domain-specific preprocessing, with one approach by Dirkson \textit{et al} \shortcite{dirkson-etal-2019-lexical} for lexical normalization of medical social media data achieving a 0.8\% reduction in out-of-vocabulary terms on spell correction tasks in cancer-related patient forums.

\section{Methods}
Our models were built using the TensorFlow implementation of BERT with cTPU on Google Colab, loosely based on code published by Dharmendra Choudhary \cite{Choudhary}. Our baseline model uses BERT\textsubscript{LARGE}  with a single additional affine layer and sigmoid for the ADR classification task with hyperparameters selected from the SMM4H 2019 winning team of Chen \textit{et al} \shortcite{chen-etal-2019-hitsz}. Three BERT-based models were then implemented with identical but alternative hyperparameters selected relative to our baseline model:
\begin{compactitem} 
  \item BERT\textsubscript{LARGE}, uncased
  \item BioBERT; initialized from BERT\textsubscript{BASE} and pre-trained on biomedical corpora, including PubMed abstracts (4.5B words) and Pubmed Central/PMC full-text articles (13.5B words)
  \item Bio + clinicalBERT (known from here onwards just as ClinicalBERT); initialized from BioBERT, then pre-trained on 2M clinical notes
\end{compactitem}

Our preprocessing approach was inspired by the lexical normalization method proposed by Dirkson \textit{et al} \shortcite{dirkson-etal-2019-lexical}. It includes several components which clean tweets and a drug normalizer module to convert brand names of drugs to their generic equivalent utilizing Dirkson’s DrugNorm method \shortcite{dirkson-2019-drugnorm}. The full list of components used in our final preprocessor are described in Table~\ref{preprocessor}.

\begin{figure*}[h!]
  \caption{Final Ensembling Approach}
  \includegraphics[width=\linewidth]{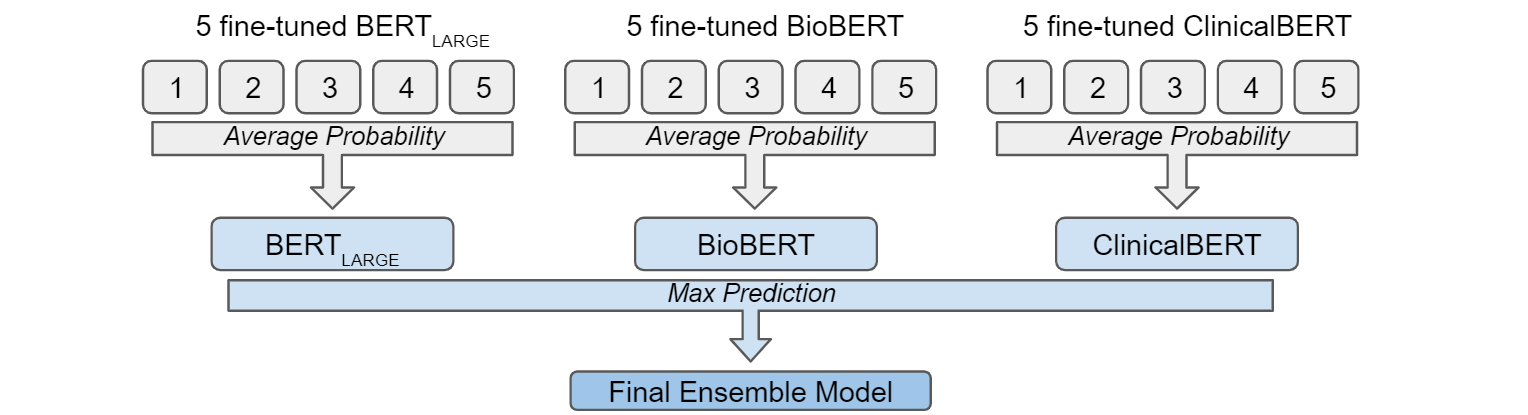}
  \label{model}
\end{figure*}

Notably, despite the success of the spell corrector component in reducing out-of-vocabulary terms in previous research, we decided to exclude it from our model due to our determination that an unacceptable number of erroneous alterations to the text were resulting from its use (see Table~\ref{spellcorrect} for examples). 

\begin{table}[t!]
\small
\caption{\label{spellcorrect}  Examples of Erroneous Alterations Due to Preprocessor Spell Corrector }
\centering
\begin{tabular}{ |l|l|l| } 
 \hline
  \bf Original & \bf Altered & \bf Likely Intended Meaning\\
 \hline
  gerd & germs & Gastroesophageal reflux disease\\
  imigran & migraine & Brand name for Sumatriptan\\
  lovan & prozac & Brand name for Fluoxetine\\
 \hline
\end{tabular}
\end{table}

We split our training set 80/20 into training and dev sets for the purposes of fine-tuning our models. Due to the model variability resulting from each fine-tuning run of the training set (described further in the Results \& Discussion section), we ran five models for each model type. We then applied an ensembling method that included averaging the predicted probabilities across runs, then using a maximum positive probability ensemble technique, whereby we predicted the positive class for a tweet if any of the three included models predicted the positive class (see Figure \ref{model}).  

Our final approach was evaluated by fine-tuning our models on the full training set (25,678 tweets) and then evaluating these on a test dataset (4,575 tweets) on the post-evaluation competition platform \cite{SMM4H'19}. Key performance metrics are consistent with the SMM4H 2019 competition: $F_1$-score, precision, and recall. However, we prioritized optimizing recall over precision for reasons described in our Introduction section. 

\section{Results}

\subsection{Baseline models}

Our baseline model, using the SMM4H winning team’s model parameters but without their corpus for retraining, reported as "BERT\_noRetrained"  \cite{chen-etal-2019-hitsz}, struggled with convergence for roughly 30\% of the models trained, resulting in $F_1$-scores of 0, given that it predicted the negative class during evaluation for every test example.  We therefore ran the model multiple times and only included the results of the first five models with non-zero scores as our baseline. The results of the five converging baseline models on the test set and their average are presented in Table~\ref{baseline}. Notably, the $F_1$-score is nearly identical to that reported by Chen \textit{et al} \shortcite{chen-etal-2019-hitsz}. Due to this convergence issue, we decreased our learning rate from 5e-5 to 2e-5 for our subsequent models.  

\begin{table}
\centering
\small
\begin{threeparttable}
\caption{\label{baseline} Average Prediction on Test set for Baseline BERT}
\begin{tabular}{ |l|c|c|c|c|c|c|c|} 

 \hline
 & \bf 1 & \bf 2 & \bf 3 & \bf 4 & \bf 5 & \bf Avg & \bf Chen\tnote{1}\\ \hline
$F_1$& 0.59 & 0.63 & 0.62 & 0.63 & 0.62 & \bf 0.618 & 0.618\\
P& 0.64 & 0.66 & 0.64 & 0.66 & 0.67 & \bf 0.654 & 0.646\\
R& 0.55 & 0.61 & 0.61 & 0.60 & 0.57 & \bf 0.587 & 0.593\\ \hline
\end{tabular}
\begin{tablenotes}
\footnotesize
\item[1] based on Chen 2019's dev set 
\end{tablenotes}
\end{threeparttable}
\end{table}

\begin{figure}[h!]
 \centering
  \caption{Training Loss After Each Epoch for 5 Runs on BioBERT with Preprocessor}
  \includegraphics[width=\linewidth]{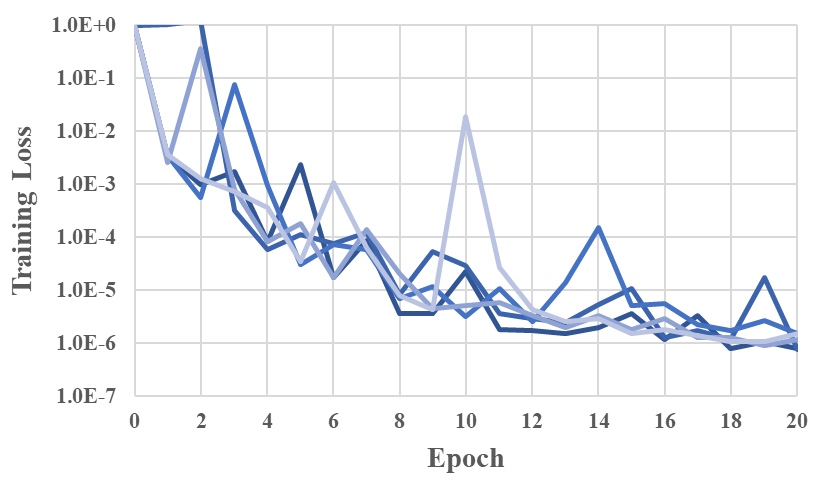}
  \label{fig:epoch}
\end{figure}

Additionally, we discovered that the number of epochs set at 20 was unnecessary for our models, likely due to the relatively small size of our dataset. While the loss after the first epoch ranged between 0.001 and 1.0, by the fourth epoch our training loss would consistently drop below 0.001 (see Figure \ref{fig:epoch}). We consequently reduced this hyperparameter to 8 to reducing training time.

\begin{table*}
\small
\begin{threeparttable}
\caption{\label{variability} Standard Deviations from 5 Runs on Scenarios Aimed to Reduce Variability in Performance}
\begin{tabular}{|p{0.5cm}|p{5cm}|p{1.9cm}|p{1.9cm}|p{1.9cm}|p{1.9cm}|}
\hline
\multicolumn{1}{|r|}{Index} & \multicolumn{1}{c|}{Scenario} & \multicolumn{2}{c|}{BERT \textsubscript{LARGE}\tnote{1}}
& \multicolumn{2}{c|}{BioBERT}\\
\cline{3-6}
\multicolumn{1}{|c|}{ }&  & $F_1$ StDev & Recall StDev &$F_1$ StDev & Recall StDev \\
\hline
0 & Original & 0.014 & 0.017 & 0.007 & 0.015\\
1 & Training set with 3x positive examples & 0.010 & 0.024 & 0.011 & 0.011\\
2 & Batch size = 128 & 0.013 & 0.022 & \cellcolor[HTML]{CFE2F3} 0.004 & 0.011\\
3 & Loss with 2x positive weights & \cellcolor[HTML]{CFE2F3} 0.007 & 0.013 & 0.011 & 0.017\\
4 & Dropout = 0.2 & 0.012 & 0.022 & \cellcolor[HTML]{CFE2F3} 0.007 & \cellcolor[HTML]{CFE2F3} 0.006\\
5 & 1\&2 combined & \cellcolor[HTML]{CFE2F3} 0.007 & 0.009 & 0.012 & 0.013\\
6 & 1\&2\&3\&4: All combined & 0.008 & 0.011 & 0.007 & 0.020\\
\hline
\end{tabular}
\begin{tablenotes}
\footnotesize
\item[1] This version of BERT\textsubscript{LARGE} also included additional preprocessing which was not ultimately included in our final models 
\end{tablenotes}
\end{threeparttable}
\end{table*}

\subsection{Model variability analysis}

We hypothesized that the variability in predictions observed as part of our baseline model were due either to the relatively small training set or due to the general rarity of positive examples in our corpus. Four alternative model specifications, and some combinations of these four alternatives, were evaluated to determine if we could minimize this variability. These four changes included:
\begin{enumerate}[nolistsep]
  \item Altering the training set by making two additional copies of each positive-label example (i.e. those labeled as an ADR)
  \item Increasing the batch size from 32 to 128
  \item Adjusting our loss function to weigh positive weights three times higher than negative weights
  \item Increasing the dropout rate from 0.1 to 0.2
\end{enumerate}

Each scenario was run 5 times on our dev set with the output metric being the standard deviation for both $F_1$-score and recall. These were run over both BERT\textsubscript{LARGE} and BioBERT models and presented in Table~\ref{variability}.

This analysis did not uncover any consistent themes on approaches that could reduce the variability observed within the model predictions. Furthermore, given that no other model outputs showed cause for concern, we accepted this variability as a consequence of our particular task and managed this issue by taking the average prediction from five model runs for each of our final BERT models. 

\begin{figure}[h!]
 \centering
  \caption{Sources of False Positives and True Positives: Max Ensemble on Dev Set with Preprocessor}
  \includegraphics[width=\linewidth]{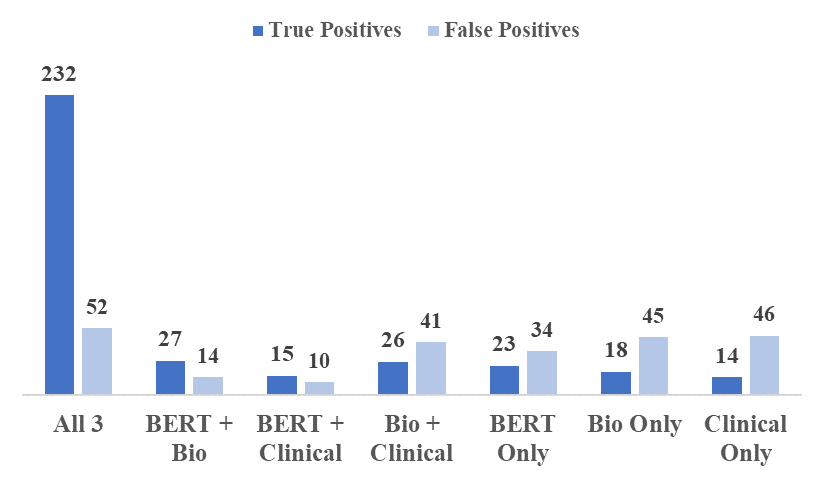}
  \label{tp_fp}
\end{figure}

\begin{table*}[t!]
\begin{center}
\caption{\label{results}Average Prediction for BERT\textsubscript{LARGE}, BioBert, ClinicalBert and Max Ensemble Results with and without Preprocessor}
\small
\begin{tabular}{|p{2.5cm}|p{2.8cm}|p{2.8cm}|p{2.8cm}|p{2.8cm}|}
 \hline
  \bf \textit{No preprocessor} & \bf BERT & \bf BioBERT & \bf ClinicalBERT & \bf Max Ensemble\\
  \rowcolor[HTML]{CFE2F3} $F_1$-score & 0.6446 & 0.5915 & 0.5809 & 0.6378\\
  Precision & 0.6695 & 0.6200 & 0.6180 & 0.5663\\
  \rowcolor[HTML]{CFE2F3}Recall & 0.6214 & 0.5655 & 0.5479 & 0.7300\\ \hline
  \bf \textit{Preprocessor} & \bf BERT PP & \bf BioBERT PP & \bf ClinicalBERT PP & \bf Max Ensemble PP\\
  \rowcolor[HTML]{CFE2F3}$F_1$-score & 0.6475 & 0.6153 & 0.6212 & 0.6681\\
  Precision & 0.6907 & 0.6577 & 0.6360 & 0.5900\\
  \rowcolor[HTML]{CFE2F3}Recall & 0.6097 & 0.5793 & 0.6076 & 0.7700\\
 \hline
\end{tabular}
\end{center}
\end{table*}

\subsection{Final models}

The average prediction results, based on five runs of each model, for each of our six model types (BERT\textsubscript{LARGE}, BioBERT, ClinicalBERT with and without our preprocessor) are presented in Table~\ref{results}. BERT\textsubscript{LARGE} outperformed BioBERT and ClinicalBERT on all performance metrics when no preprocessor was applied with $F_1$-scores of 0.65, 0.59, and 0.58 respectively. However, the difference in performance lessened considerably when the preprocessor was applied, with ClinicalBERT in particular observing improved predictions, with $F_1$-score increasing from 0.58 to 0.62 and recall increasing from 0.55 to 0.61.  

Despite the poorer performance of BioBERT and ClinicalBERT, ensemble methods demonstrate that these alternative representations capture additional information which BERT\textsubscript{LARGE} does not. This is demonstrated in Figure \ref{tp_fp} where 16.3\% of true positives in our dev set with preprocessor ensemble were captured by either BioBERT or ClinicalBERT but not BERT\textsubscript{LARGE}. Our best performing model resulted from ensembling the three BERT models with preprocessor as seen in Table~\ref{results} with $F_1$-score increasing by 0.02 and recall increasing by a substantial 0.16 points relative to our BERT\textsubscript{LARGE} model. 

The ClinicalBERT and BioBERT models contributed a high percentage of false positives relative to BERT\textsubscript{LARGE} (Figure~\ref{tp_fp}). In an attempt to reduce the number of false positives, we experimented with threshold tuning. We found that implementing a threshold of 0.6 for the BioBERT and ClinicalBERT models and 0.5 for the BERT\textsubscript{LARGE} model resulted in a higher $F_1$-score and higher recall on our training data. However, when we evaluated the models on our test data, we found that a threshold of 0.5 for all models performed slightly better in terms of $F_1$-score and recall. We believe the threshold tuning resulted in overfitting the model to the training data, thus we used the standard threshold of 0.5 for all models in the final ensemble.

\section{Discussion}

Our research suggests that ensembling various BERT implementations (general domain BERT\textsubscript{LARGE} and domain-specific BioBERT and ClinicalBERT) and applying a simple domain-specific preprocessing technique can improve predictions for ADR classification in social media. Our best ensemble model has a recall of 0.770 compared to baseline of 0.587 and $F_1$-score of 0.668 compared to baseline of 0.618. This model outperformed the best results on both metrics presented at the August 2019 meeting of SMM4H as well as exceeded all results submitted on the competition platform since the meeting in post-evaluation (as of March 21, 2020). 

Notably, our BERT\textsubscript{LARGE} model with preprocessor performed as well as the SMM4H 2019 winning team's model, which involved retraining BERT\textsubscript{LARGE} on a sizable Twitter corpus. This suggests that simple preprocessing techniques are still valuable to implement in the context of state-of-the-art representation models such as BERT. Of all components in our preprocessor, we believe the drug normalizer, which translates brand names of drugs into generic names, is the most important. After examining the outputs of models with no preprocessor, we noticed that many of the false negative examples contained the brand name of a drug, rather than the generic name. The generic names of drugs often contain similar suffixes for drugs in the same family, so when the BERT tokenizer is applied it is able to create more meaningful tokens from the generic drug names. For example, quetiapine (que \#\#tia \#\#pine) and olanzapine (o \#\#lan \#\#za \#\#pine) both share the same last token (\#\#pine), while their brand name equivalents, Seroquel (se \#\#ro \#\#quel) and Zyprexa (z \#\#y \#\#p \#\#re \#\#xa), have no tokens in common. We hypothesize this ability to create more meaningful, generalizable tokens is why the addition of the drug normalizer improved the $F_1$-score and recall for models pretrained on biomedical corpora (BioBERT and ClinicalBERT) by 2.4\% - 10.9\%. While the preprocessor did not have a similar effect on $F_1$-score and recall for BERT\textsubscript{LARGE}, this is expected as brand names of drugs are often more commonly used outside of biomedical corpora.

Despite our efforts to identify a way to minimize the variability in model predictions, each of our final models continue to have a standard deviation of approximately 0.01 which we attribute to the skewness of the classification task at hand and the limited amount of training data available. It is likely that other researchers who developed BERT-based models for this task may have suffered from similar challenges as this variance in accuracy is a known issue for BERT models trained on small datasets \cite{devlin-etal-2019-bert}. It was due to this variability that our ensembling approach begins by taking an average prediction of 5 trained models for each of our three BERT models, which effectively operates as a majority vote for each classification. Ensembling according to the max prediction across the resulting BERT\textsubscript{LARGE}, BioBERT and ClinicalBERT ensembled models ensured we maximized recall by taking information distinctly gleaned from each of these three model.

\section{Conclusion}

This study aimed to develop a state-of-the-art adverse drug reaction (ADR) detection deep learning model for social media. Our contribution has been the development of a BERT-based deep learning model combining improved domain-specific preprocessing and ensembling which has resulted in state-of-the-art results for this ADR classification task. This study also provided a clear illustration of the variability that can be expected when fine-tuning BERT models on small datasets and some practical considerations for how to overcome this variability through averaging. Despite our success in developing this model, we identified two main limitations associated with this work in the context of its ideal use case which is the real time monitoring of social media platforms for ADRs. Firstly, such a large ensemble model may not be realistic to put into production and indeed alternative methods such as simpler single-layer BiLSTM distilled from BERT-based models \cite{tang} may be more suited to a real-time task. Secondly, this classification task only represents the first task that should be automated, with ADR mention extraction and normalization also required. The performance of existing models to perform these downstream tasks have considerable room for improvement \cite{weissenbacher-etal-2019-overview} and should be the focus of further research.

\bibliography{adr_twitter}
\bibliographystyle{acl2018}

\end{document}